\documentclass[conference]{IEEEtran}
\IEEEoverridecommandlockouts

\usepackage{cite}
\usepackage{amsmath,amssymb,amsfonts}
\usepackage{algorithmic}
\usepackage{graphicx}
\usepackage{textcomp}
\usepackage{xcolor}
\usepackage{booktabs}
\usepackage{multirow} 
\def\BibTeX{{\rm B\kern-.05em{\sc i\kern-.025em b}\kern-.08em
    T\kern-.1667em\lower.7ex\hbox{E}\kern-.125emX}}
\begin{document}

\title{TrailBlazer-R: Speculative Decoding for Autoregressive Vision-Language-Action Models in Robotics
}
\author{
    Haibo Hu$^{1}$\quad Weilan Wang\textsuperscript{\rm 2}\thanks{Corresponding author: weilawang2-c@my.cityu.edu.hk}\quad Qiao Li\textsuperscript{\rm 2}\quad Nan Guan$^{1}$\quad  Chun
    Jason Xue$^{2}$ \\
    $^{1}$City University of Hong Kong\\
    $^{2}$Mohamed bin Zayed University of Artificial Intelligence
}

\maketitle

\begin{abstract}
Low-latency inference is essential for deploying autoregressive vision-language-action (VLA) models in robotic manipulation. However, existing acceleration methods often reduce computation without fully addressing the serial dependencies of action generation. We introduce TrailBlazer-R, an approximate speculative decoding framework whose lightweight drafters chart candidate action paths ahead of the frozen target, creating opportunities for parallel inference. We simplify complex seven-dimensional action prediction into three focused behavioral tasks: translational continuity, rotational continuity, and grasp timing. This decomposition enables accurate proposals from lightweight, non-autoregressive drafters trained on target-generated actions. Speculative parallel generation then brings these independently generated components together for joint evaluation in one causal target-model forward pass. Controller-aware acceptance applies motion tolerances in physical units and requires gripper-state agreement, providing interpretable criteria for accepting complete actions while maintaining comparable task performance. Experiments with WorldVLA and RynnVLA-002 achieve $1.42$--$1.44\times$ task-completion speedups across four LIBERO suites on NVIDIA L20 servers, within 0.2 percentage points of the native policies' success rates on average. Deployment on Jetson AGX Orin 64GB further achieves $1.42$--$1.52\times$ task-completion speedups on selected LIBERO-Spatial tasks. our code is https://github.com/DanileHaibo/Trailblazer-R
\end{abstract}


\section{Introduction}
\begin{figure}[!t]
\centering
\includegraphics[width=0.5\textwidth, trim=0cm 0cm 0cm 0cm, clip]{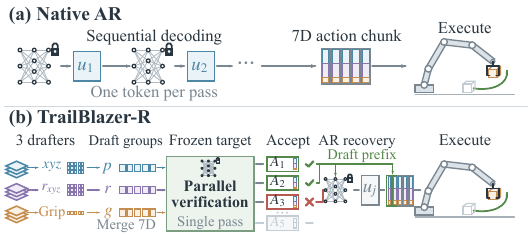}
\caption{Robotic VLA decoding: (a) native autoregressive generation; (b) TrailBlazer-R with parallel generation}
\label{fig:introduction}
\end{figure}

Vision-language-action (VLA) models advance generalist robotic manipulation by mapping visual observations and language instructions into actions \cite{b1,b2,b3}. Autoregressive (AR) policies, including OpenVLA and WorldVLA, are widely adopted for their unified token formulation, compatibility with pretrained vision-language backbones, and modeling of action-token dependencies \cite{b2,b4}. However, sequential generation repeatedly invokes a large backbone, increasing latency for longer action chunks and delaying responses to new observations (Fig.~\ref{fig:introduction}(a)) \cite{b5,b6}. Existing acceleration techniques include parallel fixed-point decoding \cite{b5}, consistency distillation and early termination of decoding iterations \cite{ceed}, layer-wise early exit \cite{b17}, and visual-token pruning or caching \cite{pruner,b18}. Approximate iterative decoding, early exit, and distillation can shift the pretrained policy's output distribution, potentially improving or degrading task performance \cite{ceed,free,raee}. Visual-token optimizations retain sequential action dependencies, limiting gains when decoding dominates latency \cite{pruner,b18}. These limitations motivate a strategy that reduces serial computation while retaining the target architecture and learned parameters, with the goal of preserving closed-loop task performance.

Speculative decoding accelerates autoregressive generation through lightweight drafting and parallel target verification, accepting multiple tokens per forward pass while preserving the target distribution under exact verification \cite{b9,b11,b12}. Applying it to robotic VLA policies presents two challenges: heterogeneous seven-dimensional actions are difficult for lightweight drafters to predict accurately and efficiently, and token-bin tolerances do not specify translation and rotation tolerances in controller units, gripper-state consistency, or joint acceptance of complete actions. We find that specializing drafters for translation, rotation, and gripper control improves accuracy over a monolithic drafter, simplifying joint prediction into narrower objectives that enable efficient drafting. Their separate outputs are assembled into complete action sequences for joint parallel target verification, naturally combining specialized prediction with unified assessment within a shared action context. We also observe that these components have distinct physical properties and different effects on task success when prediction errors occur, suggesting that acceptance can be adapted to each component's control semantics and error sensitivity.

Motivated by these challenges and findings, we propose TrailBlazer-R, an approximate speculative decoding framework tailored to robotic VLA inference. TrailBlazer-R simplifies the complex seven-dimensional drafting problem by decomposing it into three specialized prediction tasks: translational continuity, rotational continuity, and grasp timing, each handled by a lightweight non-autoregressive drafter. It then reassembles these independently generated components into complete action sequences and exploits speculative decoding's parallel generation to jointly evaluate them through a single frozen target-model forward pass. Building on this verification, TrailBlazer-R introduces a controller-aware acceptance mechanism that reflects the physical properties and error sensitivities of robotic actions. It applies separate translation and rotation tolerances in controller units, requires binary gripper-state consistency, and accepts each complete seven-dimensional action only when all component criteria are satisfied.

Our main contributions are summarized as follows:
\begin{itemize}
    \item We propose TrailBlazer-R, which simplifies seven-dimensional action drafting through three specialized non-autoregressive drafters for translation, rotation, and gripper control, trained on target-generated actions to align their predictions with the target policy.
    \item We develop joint parallel generation and controller-aware acceptance to unify independently drafted components, evaluating complete actions using component-specific physical tolerances and binary gripper-state consistency.
\item We evaluate TrailBlazer-R on two AR VLA models across four LIBERO suites, achieving $1.42$--$1.44\times$ task-completion speedups on NVIDIA L20 servers with comparable average task success, reaching $1.56\times$ on LIBERO-Spatial. On Jetson AGX Orin 64GB, TrailBlazer-R achieves $1.42$--$1.52\times$ average task-completion speedups.
\end{itemize}

\section{Background and Motivation}

\subsection{Background}
Existing VLA acceleration methods optimize different stages of inference. PD-VLA reformulates autoregressive action generation as parallel fixed-point iterations \cite{b5}. CEED-VLA combines consistency distillation with early termination to reduce decoding iterations \cite{ceed}. DeeR-VLA selects intermediate backbone exits based on action consistency \cite{b17}. VLA-Pruner prunes visual tokens using semantic, action, and temporal relevance \cite{pruner}, while VLA-Cache reuses minimally changed visual representations and recomputes task-relevant information \cite{b18}. KERV integrates Kalman-filter-based action compensation with kinematics-guided dynamic acceptance thresholds for speculative decoding \cite{Kerv}. TrailBlazer-R combines three specialized non-autoregressive drafters with joint target verification of complete actions using controller-space tolerances, retaining the target architecture and learned parameters while maintaining comparable closed-loop task performance. As an auxiliary decoding framework, it can in principle complement compatible model optimizations through target-specific drafter adaptation.

\begin{figure}[!t]
\centering
\includegraphics[width=0.5\textwidth, trim=0cm 0.6cm 0cm 0cm, clip]{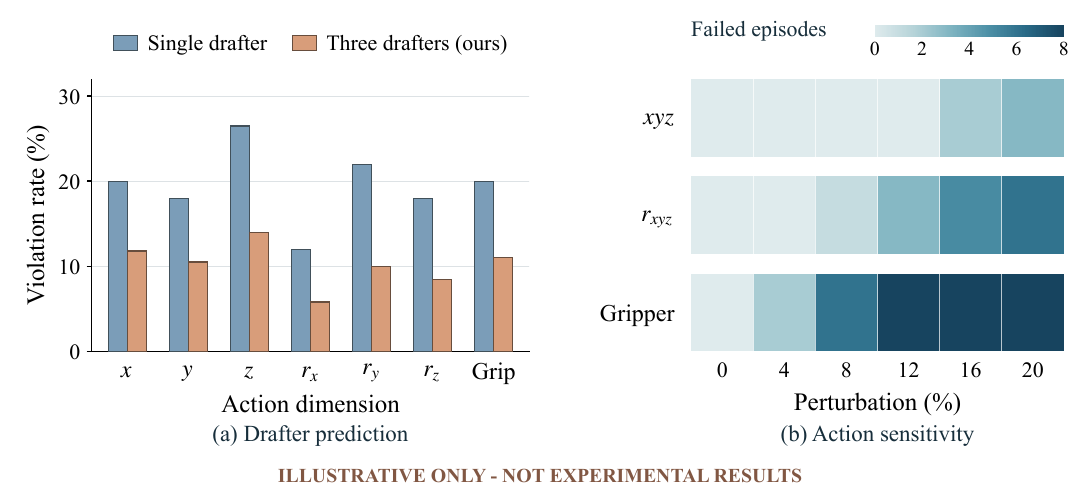}
\caption{Motivation for TrailBlazer-R: (a) violations with single versus three specialized drafters; (b) component-specific perturbation sensitivity, with darker colors indicating more failed episodes.}
\label{fig:motivation1}
\end{figure}

\begin{figure*}
    \centering
    \includegraphics[width=0.9\linewidth]{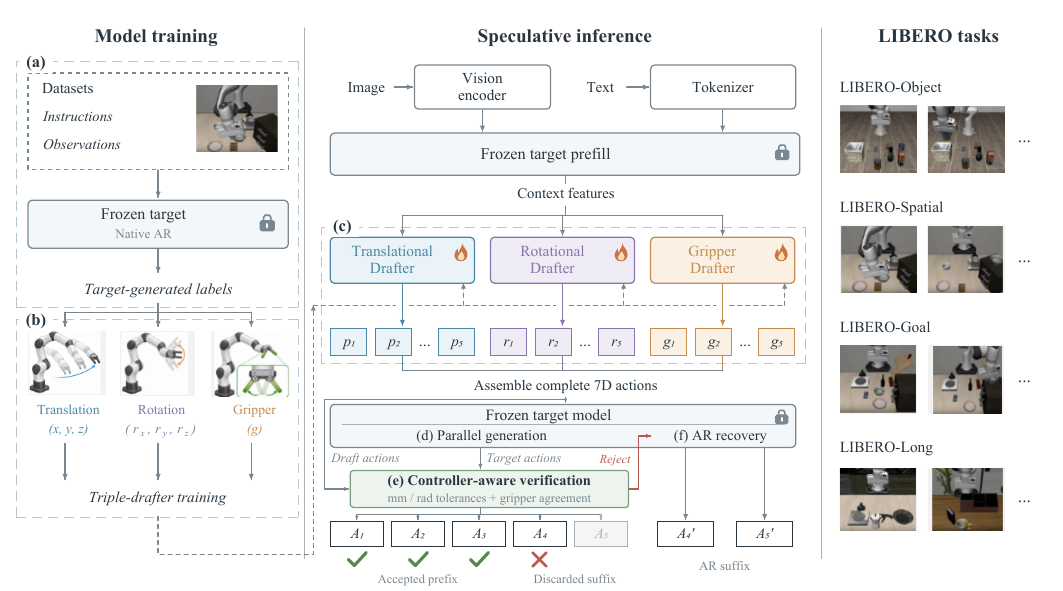}
\caption{Overview of TrailBlazer-R: (a) collecting target-generated supervision; (b) training three specialized drafters; (c) predicting action components in parallel; (d) generating target predictions conditioned on the assembled proposals; (e) jointly verifying complete actions using controller-aware criteria; and (f) regenerating the rejected suffix autoregressively.}
    \label{fig:overview}
\end{figure*}

\subsection{Motivation}

We compare a single seven-dimensional drafter with three specialized drafters using a common architecture and identical target-generated data. Fig.~\ref{fig:motivation1}(a) reports episode-weighted violation rates against independent native AR actions: per-axis translation errors above 1~mm, rotation errors above 0.01~rad, or gripper-state disagreement with binary states defined by $\mathrm{command}>0$. These thresholds match the default gate in Table~\ref{tab:gate_sensitivity}. Fewer motion violations suggest that shallow drafters, whose depth is limited by inference budgets, learn component-specific continuity more readily than heterogeneous seven-dimensional actions. Fig.~\ref{fig:motivation1}(b) reports failed-episode counts under perturbations expressed as percentages of each component's controller-space action range, using 80 episodes in total; darker colors indicate more failures. Gripper perturbations appear most consequential, followed by rotation, reflecting grasp/release and alignment requirements, while translation permits geometric tolerance. Thus, action discrepancies require component-specific physical interpretation rather than uniform magnitude-based criteria.

\section{Methodology}

\subsection{Problem Formulation}

Given visual-language context and robot state $c_t$, an autoregressive VLA predicts an $H$-step action chunk, where each action $\mathbf{a}_{t,h}=(\Delta\mathbf{p}_{t,h},\Delta\mathbf{r}_{t,h},g_{t,h})$ comprises three translation dimensions, three rotation dimensions, and a gripper command.

We measure wall-clock task completion time, including inference, environment interaction, control execution, and runtime overhead. This captures differences in execution length: faster individual predictions may still require more steps to complete a task. Let $\theta$ denote the drafter and acceptance configurations, $\bar{T}_{\mathrm{task}}(\theta)$ the mean completion time over successful episodes, and $S(\theta)$ the success rate over all attempted episodes. Our objective is:
\begin{equation}
    \min_{\theta}\;\bar{T}_{\mathrm{task}}(\theta)
    \quad\text{s.t.}\quad
    S(\theta)\geq S_{\mathrm{target}}-\varepsilon,
\end{equation}
where $S_{\mathrm{target}}$ is the original policy's success rate and $\varepsilon\geq0$ is the permitted reduction. All comparisons use matched hardware and evaluation conditions.
\subsection{Triple Drafting for Robotic Actions}

\begin{figure}
    \centering
    \includegraphics[width=0.40\linewidth,  trim=2.5cm 0.6cm 2.5cm 0cm]{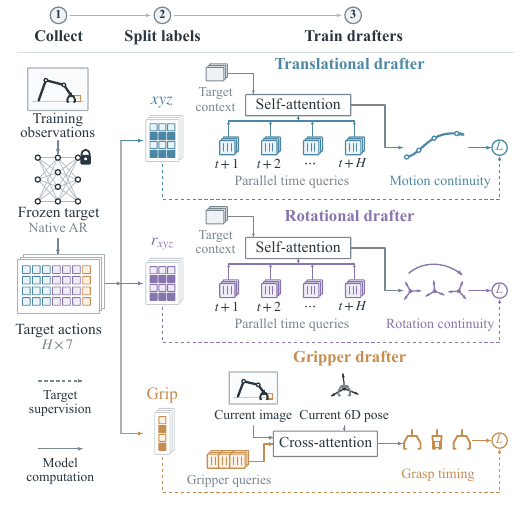}
    \caption{Triple-drafter training with factorized action labels generated by the frozen target}
    \label{fig:train}
\end{figure}

This subsection describes the model-training pipeline on the left of Fig.~\ref{fig:overview} and the three drafter branches in its speculative-inference path. Fig.~\ref{fig:train} details the training procedure in three stages: collecting target outputs, splitting action labels, and training specialized drafters.

We train a separate set of three lightweight drafters for each target VLA. During collection, the frozen target processes visual observations, language instructions, and robot states from the training split through native autoregressive inference. We record its generated action chunks as supervision, rather than using the dataset's ground-truth actions. These labels encourage the drafters to approximate the behavior of the specific target that will subsequently verify their proposals, as shown in Fig.~\ref{fig:overview}(a).

Each recorded $H\times7$ chunk, where $H$ is the prediction horizon, is split into temporally aligned translation $(x,y,z)$, rotation $(r_x,r_y,r_z)$, and gripper labels. Let $b\in\{p,r,g\}$ index translation, rotation, and gripper branches, respectively, with dimension counts $d_p=d_r=3$ and $d_g=1$. The three branches maintain independent parameters and learn translational continuity, rotational continuity, and grasp timing, respectively, from their assigned labels (Fig.~\ref{fig:overview}(b)).

The drafters reuse visual-language features from one target-model prefill and the current robot state, projecting them into branch-specific context memory $M_b$. For timestep $h\in\{1,\ldots,H\}$ and branch dimension $j\in\{1,\ldots,d_b\}$, the query is:
\begin{equation}
    q^{(b)}_{h,j}
    =
    e^{(b)}_{\mathrm{time},h}
    +
    e^{(b)}_{\mathrm{dim},j}.
\end{equation}
Here, $e^{(b)}_{\mathrm{time},h}$ and $e^{(b)}_{\mathrm{dim},j}$ are learned timestep and dimension embedding vectors of the same dimension. Stacking all $H d_b$ queries forms $Q_b$, representing every future prediction position without sequential token dependencies.

The translational and rotational drafters use self-attention across queries to capture temporal motion dependencies, while cross-attention conditions predictions on observations. Each branch thus focuses its limited capacity on positional evolution or rotational coherence. The gripper branch cross-attends to the current visual context and observed six-dimensional end-effector pose to model object relationships and grasp/release timing. This pose comes from the current observation, not from the other drafters' predicted trajectories.

Branch-specific losses compare predictions with their target-generated labels, following the dashed supervision paths in Fig.~\ref{fig:train}. Only drafter parameters are updated; the target remains frozen. At inference, each Transformer produces logits for all queries in one non-autoregressive pass:
\begin{equation}
    \widetilde{k}^{(b)}_{h,j}
    =
    \arg\max_k
    \left[D_{\phi_b}(Q_b,M_b)\right]_{h,j,k}.
\end{equation}
Here, $D_{\phi_b}$ denotes branch $b$'s Transformer with trainable parameters $\phi_b$; its output entry $[\,\cdot\,]_{h,j,k}$ is the logit for target action class $k$ at position $(h,j)$, and $\widetilde{k}^{(b)}_{h,j}$ is the selected class.

The three outputs are assembled into an $H\times7$ candidate chunk and serialized according to the target's action-token protocol, producing a complete proposal for joint parallel generation (Fig.~\ref{fig:overview}(c)).

\subsection{Joint Parallel Generation}

This stage corresponds to the action assembly and the computation inside the ``Parallel generation'' block in the central inference pipeline of Figure~\ref{fig:overview}(d). We align the three drafters' outputs by timestep, assemble complete seven-dimensional actions, and serialize them into $\widetilde{\mathbf{k}}=(\widetilde{k}_1,\ldots,\widetilde{k}_L)$ following the target's native action-token order.

In ordinary autoregressive decoding, a forward pass produces a next-token distribution, from which a token is selected before evaluating the following position. Speculative decoding supplies candidate tokens in advance, making every proposed prefix available without waiting for sequential generation \cite{b9}. The Transformer can therefore process all candidate positions simultaneously within each layer, while causal masking preserves the original conditioning order. For greedy target decoding, the predictions are:
\begin{equation}
    \widehat{k}_i
    =
    \arg\max_k
    p_{\mathrm{T}}\!\left(
    k\mid c_t,\widetilde{k}_{<i}
    \right),
    \qquad i=1,\ldots,L.
\end{equation}

Using the cached context, prefill logits predict the first candidate, while the output at draft position $i$ predicts position $i+1$. One verification forward pass therefore supplies the remaining predictions in parallel. Its final position also produces a next-token distribution beyond the candidate block, usable when the full conditioning prefix is retained and further output is needed. These predictions are conditioned on the assembled draft prefixes and provide the target references for the following verification stage.

\subsection{Controller-Aware Verification and Acceptance}

This stage corresponds to ``Controller-aware acceptance'' below the frozen target model in Figure~\ref{fig:overview}(e). We convert both draft tokens and target predictions into controller-space commands using the target's native action decoder, including any inverse normalization and controller scaling. This yields translation increments in millimeters, rotation increments in radians, and gripper commands, allowing comparison through physical quantities rather than token identities.
\begin{table*}[t]
\centering
\caption{Task success and end-to-end acceleration across four LIBERO suites. Completion time is averaged over successful episodes, and speedup is measured relative to each target's native autoregressive decoding.}
\label{tab:discrete-libero}
\setlength{\tabcolsep}{8pt}
\renewcommand{\arraystretch}{1.12}
\resizebox{\textwidth}{!}{%
\begin{tabular}{lcccccccc}
\toprule
Method & Spatial & Object & Goal & Long & Avg-Task
& Avg-Time (s)$\downarrow$ & Avg-Speedup$\uparrow$ & Avg-Acceptance (\%)$\uparrow$ \\
\midrule
\multicolumn{9}{l}{\textit{Published results}} \\
OpenVLA~\cite{b2}
& 84.7 & 88.4 & 79.2 & 53.7 & 76.5 & -- & -- & -- \\
SpatialVLA~\cite{spatialvla}
& 88.2 & 89.9 & 78.6 & 55.5 & 78.1 & -- & -- & -- \\
CoT-VLA~\cite{cotvla}
& 87.5 & 91.6 & 87.6 & 69.0 & 83.9 & -- & -- & -- \\
$\pi_0$-FAST~\cite{fast}
& 96.4 & 96.8 & 88.6 & 60.2 & 85.5 & -- & -- & -- \\
\midrule
\multicolumn{9}{l}{\textit{Our paired evaluations}} \\
WorldVLA (Native)
& 86.0 & 87.4 & 80.1 & 57.5 & 77.7
& 74.8 & $1.00\times$ & -- \\
WorldVLA + TrailBlazer-R
& 85.8 & 87.7 & 79.8 & 57.2 & 77.6
& 51.9 & $1.44\times$ & 48.7 \\
\addlinespace[2pt]
RynnVLA-002 (Native)
& 94.0 & 96.4 & 95.1 & 87.4 & 93.2
& 72.5 & $1.00\times$ & -- \\
RynnVLA-002 + TrailBlazer-R
& 93.2 & 96.0 & 96.4 & 86.4 & 93.0
& 51.1 & $1.42\times$ & 45.1 \\
\bottomrule
\end{tabular}%
}
\end{table*}

\begin{table}[t]
\centering
\caption{VLA inference acceleration on the complete LIBERO-Spatial task suite}
\label{tab:vla_acceleration}
\scriptsize
\setlength{\tabcolsep}{4pt}
\renewcommand{\arraystretch}{1.05}
\begin{tabular}{@{}lrrrrrr@{}}
\toprule
 & \multicolumn{3}{c}{WorldVLA}
 & \multicolumn{3}{c}{RynnVLA-002} \\
\cmidrule(lr){2-4}\cmidrule(l){5-7}
Method
 &Time(s)$\downarrow$ & Spe.($\times$)$\uparrow$ & Succ.$\uparrow$
 & Time(s)$\downarrow$ & Spe.($\times$)$\uparrow$ & Succ.$\uparrow$ \\
\midrule
Native
 & 33.5 & 1.00 & 86.0
 & 33.2 & 1.00 & 94.0 \\
PD-VLA~\cite{b5}
 & 23.8 & 1.41 & 85.6
 & 24.1 & 1.38 & 89.0 \\
CEED-VLA~\cite{ceed}
 & 24.1 & 1.39 & 85.9
 & 24.6 & 1.35 & 87.7 \\
DeeR-VLA~\cite{b17}
 & 17.8 & 1.88 & 63.3
 & 18.0 & 1.84 & 68.8 \\
\midrule
VLA-Pruner~\cite{pruner}
 & 32.2 & 1.04 & 84.5
 & 31.6 & 1.05 & 86.7 \\
VLA-Cache~\cite{b18}
 & 32.8 & 1.02 & 50.1
 & 32.5 & 1.02 & 55.5 \\
\midrule
\textbf{TrailBlazer-R}
 & \textbf{21.5} & \textbf{1.56} & \textbf{85.8}
 & \textbf{22.3} & \textbf{1.49} & \textbf{93.2} \\
\bottomrule
\end{tabular}
\end{table}
\begin{table}[t]
\centering
\caption{Gate-threshold sensitivity of TrailBlazer-R on LIBERO-Spatial}
\label{tab:gate_sensitivity}
\scriptsize
\setlength{\tabcolsep}{4pt}
\renewcommand{\arraystretch}{1.05}
\begin{tabular}{@{}lrrrrrr@{}}
\toprule
 & \multicolumn{3}{c}{WorldVLA} & \multicolumn{3}{c}{RynnVLA-002} \\
\cmidrule(lr){2-4}\cmidrule(l){5-7}
Gate
 &Time(s)$\downarrow$ & Spe.($\times$)$\uparrow$ & Succ.$\uparrow$
 &Time(s)$\downarrow$ & Spe.($\times$)$\uparrow$ & Succ.$\uparrow$ \\
\midrule
Native
 & 33.5 & 1.00 & 86.0
 & 33.2 & 1.00 & 94.0 \\
\midrule
1\,mm / 5\,mrad
 & 31.9 & 1.05 & 86.2
 & 34.6 & \textit{0.96} & 93.8 \\
2\,mm / 10\,mrad
 & 31.3 & 1.07 & 85.9
 & 34.6 & \textit{0.96} & 94.2 \\
5\,mm / 25\,mrad
 & 29.9 & 1.12 & 85.3
 & 33.9 & \textit{0.98} & 93.8 \\
\textbf{10\,mm / 50\,mrad}
 & \textbf{21.5} & \textbf{1.56} & \textbf{85.8}
 & \textbf{22.3} & \textbf{1.49} & \textbf{93.2} \\
20\,mm / 100\,mrad
 & 14.9 & 2.25 & 71.3
 & 12.0 & 2.77 & 83.1 \\
40\,mm / 200\,mrad
 & 14.6 & 2.29 & 45.6
 & 11.6 & 2.87 & 51.6 \\
80\,mm / 400\,mrad
 & 7.9 & 4.23 & 43.2
 & 13.0 & 2.55 & 44.7 \\
\bottomrule
\end{tabular}
\end{table}

Let $(\widetilde{\mathbf{p}}_h,\widetilde{\mathbf{r}}_h,\widetilde{g}_h)$ and $(\widehat{\mathbf{p}}_h,\widehat{\mathbf{r}}_h,\widehat{g}_h)$ denote the decoded draft and target actions at step $h$. We accept a complete action only when:
\begin{equation}
\begin{aligned}
    v_h ={}&
    \left[\|\widetilde{\mathbf{p}}_h-\widehat{\mathbf{p}}_h\|_\infty
    \leq \epsilon_p\right] \\
    &\land
    \left[\|\widetilde{\mathbf{r}}_h-\widehat{\mathbf{r}}_h\|_\infty
    \leq \epsilon_r\right] \\
    &\land
    \left[\mathbb{I}(\widetilde{g}_h>0)
    =\mathbb{I}(\widehat{g}_h>0)\right].
\end{aligned}
\end{equation}
Here, $\epsilon_p$ and $\epsilon_r$ are configurable translation and rotation tolerances. The infinity norm requires every motion dimension to satisfy its corresponding threshold, while the gripper check requires identical binary open/closed states.

For example, decoded translations of $(1.00,1.20,2.10)$ and $(1.01,1.20,2.10)$~mm differ by only $0.01$~mm along $x$. Their translation check passes whenever $\epsilon_p\geq0.01$~mm, avoiding rejection solely because their token representations differ. The thresholds can be adjusted to balance acceptance and task performance.

All seven dimensions must pass before a timestep is accepted. We retain only the contiguous prefix of accepted draft actions: once any component fails, the entire corresponding action and all subsequent draft actions are discarded. The target then resumes autoregressive generation from that action boundary, conditioned on the retained prefix.

\textbf{Execution and Recovery.}
After verification, the runtime retains the prompt KV cache and states of the $K$ accepted actions, discards the rejected suffix, and updates decoding positions and input history. If all $H$ actions and the termination token pass, it returns the candidate chunk. Otherwise, the target completes the suffix greedily:
\begin{equation}
    Y_{\mathrm{out}}
    =
    \widetilde{Y}_{1:m}
    \oplus
    \operatorname{GreedySuffix}_{\mathrm{target}}
    \!\left(P,\widetilde{Y}_{1:m}\right),
\end{equation}
where $P$ is the prompt, $m$ is the accepted token count, and $\oplus$ denotes concatenation. Recovery uses the actual accepted candidates as context. When $K=0$, it starts from the cached prompt and prefill logits; if only termination fails, it completes that boundary, as shown in Fig.~\ref{fig:overview}(f). 

\section{Experiments}
\subsection{Experimental Setup}

\textbf{Hardware and target models.}
We train the drafters on a server equipped with four NVIDIA L20 GPUs and evaluate edge deployment on an NVIDIA Orin platform with 64\,GB of memory. We consider two target models: WorldVLA~\cite{b4}, a representative autoregressive VLA model, and RynnVLA-002~\cite{rynn}, a higher-performing target model. Both are evaluated using discrete autoregressive action generation. This selection allows us to assess the effectiveness of TrailBlazer-R across different baseline policy performance levels. For each target, we train a separate set of three specialized drafters using target-generated action predictions, while keeping the target model's architecture and parameters unchanged.

\textbf{Benchmark and evaluation protocol.}
We evaluate on LIBERO-Spatial, Object, Goal, and Long~\cite{libero}, each containing ten tasks. We run 50 rollouts per task using benchmark-provided initial states, totaling 2,000 rollouts per evaluation run, with success determined by LIBERO's completion criteria. Detailed comparisons and sensitivity analyses focus on LIBERO-Spatial, as spatial reasoning and precise object placement are fundamental to robotic manipulation.
\begin{table*}[htbp]
\centering
\caption{WorldVLA and RynnVLA-002 on Jetson AGX Orin}
\label{tab:jetson-spatial-both}
\small
\begin{tabular}{llrrrrcc}
\toprule
& & \multicolumn{2}{c}{Completion latency (s)}
& & & \multicolumn{2}{c}{Success} \\
\cmidrule(lr){3-4}\cmidrule(lr){7-8}
Model & Task & Native & TrailBlazer-R & Speedup & Acceptance
& Native & Accel. \\
\midrule
\multirow{5}{*}{WorldVLA}
& Between plate \& ramekin & 273.981 & 193.900 & $1.413\times$ & 49.00\% & 8/10 & 9/10 \\
& Next to ramekin & 404.849 & 283.619 & $1.427\times$ & 50.13\% & 10/10 & 10/10 \\
& Table center & 341.644 & 227.986 & $1.499\times$ & 51.81\% & 10/10 & 10/10 \\
& On cookie box & 309.955 & 237.743 & $1.304\times$ & 38.31\% & 10/10 & 10/10 \\
& Avg & 336.388 & 236.887 & $1.420\times$ & 47.79\% & 38/40 & 39/40 \\
\midrule
\multirow{5}{*}{RynnVLA-002}
& Between plate \& ramekin & 164.526 & 105.780 & $1.555\times$ & 31.30\% & 9/10 & 9/10 \\
& Next to ramekin & 191.756 & 120.818 & $1.587\times$ & 46.09\% & 10/10 & 10/10 \\
& Table center & 176.056 & 113.246 & $1.555\times$ & 44.65\% & 10/10 & 10/10 \\
& On cookie box & 154.277 & 108.179 & $1.426\times$ & 36.00\% & 9/10 & 9/10 \\
& Avg & 170.973 & 112.270 & $1.523\times$ & 40.35\% & 38/40 & 38/40 \\
\bottomrule
\end{tabular}
\end{table*}

\subsection{Main Results on LIBERO}

\textbf{Task completion time.}
Table~\ref{tab:discrete-libero} shows that TrailBlazer-R reduces average task completion time from 74.8 to 51.9~s for WorldVLA and from 72.5 to 51.1~s for RynnVLA-002, achieving $1.44\times$ and $1.42\times$ speedups with acceptance rates of 48.7\% and 45.1\%. Times cover complete successful episodes, including inference, environment interaction, and control. LIBERO-Spatial profiling shows similar per-token costs (39.0 versus 40.7~ms), comparable token workloads, and nearly identical environment costs, explaining the close completion times and approximately 1.5~s difference in cumulative inference time. Both targets retain their original architectures and parameters.

\textbf{Task performance.}
The acceleration is accompanied by modest changes in task success. WorldVLA's average success rate changes from 77.7\% to 77.6\%, with individual suite-level changes within 0.3 percentage points. For RynnVLA-002, average success changes from 93.2\% to 93.0\%. Its LIBERO-Goal success rate improves from 95.1\% to 96.4\%, while the other suites exhibit decreases of 0.4--1.0 percentage points. These results demonstrate consistent task-completion acceleration with comparable aggregate task performance across both targets. 

\subsection{Comparison with Existing VLA Acceleration Methods}

Table~\ref{tab:vla_acceleration} compares acceleration methods on the complete LIBERO-Spatial suite. TrailBlazer-R reduces task completion time from 33.5 to 21.5~s for WorldVLA and from 33.2 to 22.3~s for RynnVLA-002, achieving $1.56\times$ and $1.49\times$ speedups. Both exceed those of PD-VLA and the evaluated CEED-VLA variant without consistency distillation. WorldVLA's success rate of 85.8\% is comparable to their 85.6\% and 85.9\%, while RynnVLA-002's 93.2\% exceeds their 89.0\% and 87.7\%. Relative to native execution, TrailBlazer-R reduces success by only 0.2 and 0.8 percentage points.

DeeR-VLA achieves higher speedups of $1.88\times$ and $1.84\times$, but its adaptation to these AR policies reduces success by 22.7 and 25.2 percentage points. KERV targets single-action OpenVLA decoding and not adaptation for the multi-step action chunks evaluated here \cite{Kerv}. VLA-Pruner and VLA-Cache provide only $1.02$--$1.05\times$ speedups, consistent with sequential action generation remaining a bottleneck. TrailBlazer-R reduces serial decoding through auxiliary drafting and parallel target verification, while remaining potentially complementary to compatible visual-token pruning and caching techniques through target-specific drafter adaptation.

\subsection{Sensitivity to Acceptance Thresholds}

Table~\ref{tab:gate_sensitivity} examines translation and rotation tolerances while retaining gripper-state agreement. Strict thresholds of 1--5~mm and 5--25~mrad provide limited acceleration and can even slow RynnVLA-002. The default 10~mm / 50~mrad configuration achieves $1.56\times$ and $1.49\times$ speedups with near-native success rates. Relaxing the thresholds to 20~mm / 100~mrad increases speedups on successful episodes to $2.25\times$ and $2.77\times$, but lowers overall success to 71.3\% and 83.1\%. Further relaxation reduces success to 43.2\%--51.6\%. These results support bounded physical tolerances that balance acceleration against task performance.
\begin{figure}[htbp]
    \centering
    \includegraphics[width=1\linewidth,  trim=0cm 0cm 0cm 0cm]{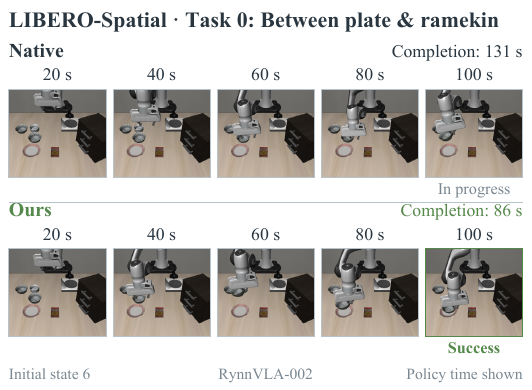}
    \vspace{-20px}
    \caption{Task visualization of native inference and TrailBlazer-R on LIBERO-Spatial Task~0 using Jetson AGX Orin.}
    \label{fig:vis}
\end{figure}

\subsection{On-Device Evaluation on Jetson AGX Orin}
We evaluate TrailBlazer-R on NVIDIA Jetson AGX Orin 64\,GB using four LIBERO-Spatial simulation tasks (IDs 0--3), with ten initial states each and 40 episodes per model and method. WorldVLA uses FP32 and RynnVLA-002 BF16; this precision difference may partly explain their absolute completion-time gap. Table~\ref{tab:jetson-spatial-both} reports completion times over successful episodes, acceptance rates, and success counts. Against same-precision native baselines, TrailBlazer-R achieves $1.304$--$1.499\times$ speedups for WorldVLA and $1.426$--$1.587\times$ for RynnVLA-002. Success remains unchanged on three tasks per model; on the remaining task, WorldVLA improves from 8/10 to 9/10, while RynnVLA-002 decreases from 9/10 to 9/10, demonstrating edge acceleration with small observed success changes.
\subsection{Qualitative Evaluation on Jetson AGX Orin}

Figure~\ref{fig:vis} compares native autoregressive inference and TrailBlazer-R using RynnVLA-002 deployed on Jetson AGX Orin 64GB. The evaluation runs in LIBERO simulation on Spatial Task~0, \textit{Between plate \& ramekin}, with both rollouts starting from initial state~6. Frames are sampled at 20-second intervals of policy time. For this episode, TrailBlazer-R completes the task in 86~s, compared with 131~s for native inference, yielding a 34.4\% reduction in completion time and a $1.52\times$ speedup. At the displayed 100-second checkpoint, native execution remains in progress, whereas TrailBlazer-R has already succeeded. This example illustrates earlier successful task completion with TrailBlazer-R on an embedded computing platform.
\section{Conclusion}

We introduced TrailBlazer-R to exploit robotic action structure for efficient autoregressive VLA decoding. By separating complex action prediction into specialized drafting tasks and reunifying their proposals through target evaluation and controller-aware acceptance, it reduces sequential computation while retaining the pretrained target architecture and parameters. Experiments demonstrate faster task completion with comparable task success. As an auxiliary decoding framework, TrailBlazer-R is also potentially complementary to compatible model compression and visual-context optimizations through target-specific drafter adaptation. This work provides a practical path toward efficient AR VLA deployment by connecting speculative execution with the structure and physical semantics of robotic actions.
\section{Content Generated by AI}
Codex was used solely as a general-purpose aid for editing and debugging.


\end{document}